\documentclass[conference]{IEEEtran}
\IEEEoverridecommandlockouts

\usepackage{booktabs} 
\usepackage{multirow}

\usepackage[bottom]{footmisc}
\usepackage{blindtext}
\usepackage{graphicx}
\usepackage[font=footnotesize]{caption,subcaption}
\usepackage{wrapfig}

\usepackage{flushend}
\usepackage{amsmath,amsfonts,amsthm,bm}
\usepackage{stfloats}
\usepackage{graphicx}
\usepackage{textcomp}
\usepackage{xcolor}
\usepackage{booktabs} 

\usepackage{amsmath}
\usepackage{rotating}
\usepackage{booktabs}
\usepackage{listings}

\usepackage{listings}

\usepackage{ragged2e} 
\usepackage{algorithm}
\usepackage[algo2e]{algorithm2e} 
\usepackage{algpseudocode}

\def\BibTeX{{\rm B\kern-.05em{\sc i\kern-.025em b}\kern-.08em
    T\kern-.1667em\lower.7ex\hbox{E}\kern-.125emX}}
\begin{document}

\title{AttnGen: Attention-Guided Saliency Learning for Interpretable Genomic Sequence Classification

}

\author{
\IEEEauthorblockN{Rayhaneh Shabani Nia}
\IEEEauthorblockA{
\textit{University of California, Davis}\\
California, USA\\
rshabaninia@ucdavis.edu}
\and
\IEEEauthorblockN{Ali Karkehabadi}
\IEEEauthorblockA{
\textit{University of California, Davis}\\
California, USA\\
akarkehabadi@ucdavis.edu}
}




\maketitle

\begin{abstract}
Deep neural networks have achieved great performance in genomic sequences; however, still is a mystery to relate their predictions to biologically meaningful patterns. In this work, we present \textbf{AttnGen}, an attention-guided training framework that embeds interpretability directly into the optimization process. AttnGen computes nucleotide-level importance through a specific attention mechanism. It then uses these scores to progressively suppress low-contribution positions during training. This encourages the model to concentrate its predictions on a compact set of informative regions and prevents distributing importance across noisy sequence elements. We evaluate AttnGen on the standardized \textit{demo\_human\_or\_worm} benchmark, consisting of binary classification over 200-nucleotide sequences. With moderate masking, AttnGen achieves a validation accuracy of 96.73\%, outperforming a conventional CNN baseline (95.83\%) while also exhibiting faster convergence and improved training stability. To examine whether the learned importance scores reflect functionally relevant signal, we perform perturbation-based analysis by removing high-saliency nucleotides. This results in a substantial accuracy drop—from 96.9\% to near chance level on a 3,000-sequence evaluation set—indicating that predictions depend on a relatively small subset of positions.

Our conducted analysis shows that masking 10--20\% of positions yields the most favorable trade-off between predictive performance and interpretability. These results indicate that attention-guided masking not only improves classification performance but also reshapes how models distribute importance across sequence positions. While this study targets the short genomic sequences, the proposed approach could be extend interpretable training strategies to more complex sequence modeling settings.
\end{abstract}


\section{Introduction}
In genomic sequence modeling, predictive accuracy alone is often insufficient. Model outputs are frequently used to guide downstream biological interpretation, such as identifying regulatory motifs or prioritizing candidate regions for experimental validation. Models that perform well but offer no insight into their decision process are difficult to trust. Early approaches such as position weight matrices (PWMs) and $k$-mer–based models provided a structured way to represent local sequence preferences~\cite{stormo2000dna}. However, these methods assume limited interaction structure and struggle when regulatory signals depend on broader context or long-range dependencies. Deep learning models address part of this limitation by learning representations directly from raw sequence data. Convolutional neural networks (CNNs), in particular, have shown strong performance across tasks such as transcription-factor binding prediction and enhancer detection~\cite{lecun2015deep, karkehabadi2024evaluating}. In several cases, these models rediscover known biological motifs without explicit supervision, indicating that they capture meaningful structure in the data. Recent work in self-supervised learning has explored alignment-based objectives to improve representation quality without requiring large labeled datasets. In particular, alignment learning has been applied in medical image segmentation to enforce consistency across different view and learn stable feature correspondences \cite{inproceedings}. This perspective is relevant to genomic sequence modeling, where only a subset of positions contributes meaningfully to prediction. Post hoc explanation methods attempt to bridge this gap. Saliency maps~\cite{simonyan2013deep}, integrated gradients~\cite{sundararajan2017axiomatic}, DeepLIFT~\cite{shrikumar2017learning}, and attention-based approaches~\cite{avsec2021effective} are commonly used to assign importance scores to individual nucleotides. However, these methods have limitations. For example, prior work has shown that certain saliency maps remain largely unchanged even when model parameters are randomized ~\cite{adebayo2018sanity}, raising concerns about whether they reflect true model behavior. If interpretability is introduced only after training, it does not affect how model forms its decisions. An alternative is to incorporate interpretability during training. Prior work has explored this idea through saliency-guided masking and consistency constraints~\cite{ross2017right, ismail2021improving, karkehabadi2024smoot, karkehabadi2024hlgm}. In a related direction, Unified Gravity Loss~\cite{karkehabadi2025unified} improves robustness by shaping the feature space during training. Despite its relevance, this direction remains relatively underexplored in genomic sequence modeling. AttnGen follows this perspective by integrating a lightweight attention mechanism that estimates nucleotide importance during the forward pass. Positions identified as less informative are progressively masked, encouraging the model to focus on a smaller set of discriminative regions while preserving necessary context.

We evaluate this approach on the standardized \texttt{demo\_human\_or\_worm} dataset from the \textit{Genomic Benchmarks} collection~\cite{gresova2023genomic}. Our goal is to study whether masking low-importance positions preserves predictive performance and whether the learned importance scores are aligned with the models predictions. To study this, we perform gradient based ablation by removing high  and low saliency nucleotides and measuring the resulting change in classification accuracy.

\section{Related Work}

\subsection{Deep Learning for Genomic Sequence Modeling}

Unlike images or natural language, genomic sequences do not exhibit clear spatial or semantic segmentation, making hand-crafted feature design both domain intensive and brittle. Deep learning addresses this challenge by enabling models to learn representations directly from raw DNA sequences without relying on predefined features. Early work demonstrated that convolutional neural networks (CNNs) can capture regulatory patterns from sequence data alone. Alipanahi et al.~\cite{alipanahi2015predicting} showed that CNNs could infer DNA- and RNA binding protein specificities without predefined motif templates, providing one of the first clear demonstrations of end-to-end learning in this domain. Around the same time, DeepSEA~\cite{zhou2015predicting} introduced a multi task convolutional framework capable of predicting chromatin effects at single nucleotide resolution, highlighting how small sequence variations can lead to measurable regulatory changes. As architectures became more expressive, attention shifted toward modeling interactions beyond local receptive fields. Regulatory elements often involve motifs separated by tens or even hundreds of nucleotides, requiring models to capture long-range dependencies. DanQ~\cite{quang2016danq} addressed this limitation by combining convolutional layers with bidirectional LSTMs, enabling motif level representations to interact across longer sequence spans. More recent work has incorporated attention mechanisms to model distal enhancer promoter interactions that may span kilobases ~\cite{avsec2021effective}. Collectively, these approaches reflect a transition from local motif detection toward modeling distributed and context dependent regulatory structure. Despite these advances, evaluating progress in genomic modeling remains challenging. Reported performance improvements often depend on preprocessing pipelines, filtering strategies, or dataset splits. Even minor implementation choices can lead to nontrivial differences in results, making it difficult to attribute gains to model design alone. The introduction of \textit{Genomic Benchmarks}~\cite{gresova2023genomic} addressed part of this issue by providing a curated collection of datasets with standardized preprocessing and baseline implementations. However, while such benchmarks improve evaluation consistency, they do not fully resolve how different training strategies—particularly those targeting interpretability behave under controlled conditions.

In this work, we adopt the \texttt{demo\_human\_or\_worm} dataset from the Genomic Benchmarks collection. The dataset contains 100{,}000 DNA sequences of length 200 and defines a balanced binary classification task. Its controlled setup allows us to study the effects of saliency-guided training without confounding variability introduced by custom data processing pipelines.

\subsection{Interpretability and Saliency-Guided Training}

Most interpretability methods operate after model training. Gradient-based saliency maps~\cite{simonyan2013deep} estimate input importance through local sensitivity, while Integrated Gradients~\cite{sundararajan2017axiomatic} and Grad-CAM~\cite{selvaraju2017grad} provide refinements intended to improve attribution quality. However, these approaches have known limitations: gradient-based explanations can be noisy and sensitive to small perturbations, while perturbation-based methods are often computationally expensive. In genomic applications, such instability is particularly problematic, as small changes in importance scores can directly affect biological interpretation.

More fundamentally, prior work has shown that certain saliency methods may produce visually plausible explanation even when model parameters are randomized~\cite{adebayo2018sanity}, raising concerns about whether these explanations reflect true model behavior or merely plausible artifacts. At the same time, integrating gradient-based saliency into the training loop is not straight forward. Computing saliency at each iteration requires additional backward passes and can introduce instability during optimization~\cite{kapishnikov2021guided}. This makes it difficult to directly incorporate attribution signals into the learning process. An alternative is to estimate importance scores in the forward pass. An attention mechanism provides such a pathway, it produces differentiable importance weights without requiring repeated gradient computations, making it more suitable for integration into the training objective. Saliency-Guided Training (SGT) ~\cite{ismail2021improving} builds on this idea by masking low importance features and enforcing consistency between original and masked predictions using a KL divergence regularizer. This approach encourages models to rely less on noisy or incidental features and more on stable, predictive structure.

We adapt this principle to genomic sequence classification in AttnGen. Instead of relying on gradient-based saliency during training, we use a lightweight attention mechanism to estimate per nucleotide importance in the forward pass. Progressive masking and KL-based consistency are retained, but reformulated for sequence data, where masking individual nucleotides introduces different structural constraint compared to masking pixels in images.

\subsection{Research Gaps and Our Contributions}

Although genomic deep learning architectures have become increasingly sophisticated, relatively little work has examined how interpretability constraints influence sequence based optimization. Existing saliency-guided approaches have largely been developed in vision settings, where masking operates over continuous pixel intensities. In contrast, genomic sequences are discrete and symbolic, and masking individual nucleotides can alter both local context and downstream representation in nontrivial ways.

As a result, it remains unclear whether saliency guided training strategies transfer effectively to genomic sequence modeling, particularly under standardized evaluation setting. In this work, we investigate this question through AttnGen, an attention-guided saliency learning framework. By combining forward-pass importance estimation with structured masking and consistency constraints, we examine whether interpretability can be incorporated into the training process in a way that directly shapes nucleotide-level importance and the resulting biological interpretations.

\section{Problem Statement}

\subsection{Task Definition}

Let $\mathcal{D} = \{(\mathbf{x}_i, y_i)\}_{i=1}^N$ denote a genomic dataset in which each $\mathbf{x}_i \in \Sigma^L$ is a DNA sequence of length $L$ over the nucleotide alphabet $\Sigma = \{A, T, G, C\}$, and $y_i \in \{0,1\}$ indicates the class label ( human versus \textit{C.~elegans}). Our goal is to learn a classifier $f_\theta: \Sigma^L \rightarrow \mathbb{R}^2$ that performs well on unseen sequences while also revealing which positions in the sequence contribute most strongly to its decisions. In addition to predictive accuracy, we examine whether the model’s outputs are supported by biologically meaningful patterns rather than superficial correlations. Many existing approaches prioritize predictive performance as the primary optimization target. Although some incorporate domain knowledge or motif constraints, standard end to end neural training typically does not distinguish between highly informative positions and those that contribute little signal. In practice, genomic sequences are not uniformly informative. Certain regions contain regulatory motif or conserved subsequences, whereas others introduce redundancy or noise. A central question therefore emerges: can the training process itself encourage the model to concentrate on discriminative positions, instead of relying on diffuse or dataset specific cues?

\subsection{Challenges and Research Questions}

Convolutional sequence models provide strong classification accuracy, yet they offer limited transparency regarding which nucleotides drive predictions. Attribution techniques can estimate importance score, but these are usually computed after training and do not alter how representations are formed. This separation between learning and explanation creates a mismatch: the model may rely on feature that appear weak or unstable under post-hoc analysis, making biological interpretation uncertain. Another difficulty arises from positional and compositional biases present in many genomic datasets. Models may exploit such biases to achieve high training accuracy without learning relationships that generalize beyond the dataset at hand. Additionally, incorporating gradient-based saliency directly into optimization introduces computational overhead and can amplify instability during backpropagation ~\cite{kapishnikov2021guided}. These considerations lead us to three guiding questions. We first examine whether attention mechanisms can approximate gradient derived saliency in a way that remains stable during training. We then study how masking sequence positions based on learned importance influences model focus and predictive behavior. Finally, we investigate how different masking intensities affect the balance between accuracy, interpretability, and robustness in nucleotide-level classification.

\subsection{Approach Overview}

We propose \textbf{AttnGen}, an attention-guided saliency learning framework that incorporates interpretability constraints into the optimization process. Instead of computing gradient based attribution at every iteration—which typically requires additional backward passes—AttnGen introduces a lightweight attention module that produces per-nucleotide importance scores during the forward computation. This design avoids repeated gradient evaluations and reduces the computational cost relative to gradient-based saliency integration.

The model then applies progressive masking to positions assignd low importance. Rather than enforcing hard sparsity, this masking encourages the network to rely more consistently on discriminative regions of the sequence. To maintain stable predictions, a Kullback–Leibler divergence term aligns the output distributions of original and masked input. This regularization penalizes large predictive shifts when low-importance positions are removed, thereby reinforcing the relevance of retained nucleotides. Our focus is to analyze how this attention-driven masking strategy shapes model behavior under controlled experimental conditions and whether it improves the alignment between predictive performance and identifiable sequence regions.

\section{Methodology}

\subsection{Model Architecture}

The base classifier $f_\theta$ in AttnGen is a 1D convolutional network designed for genomic sequence inputs. An embedding layer first maps discrete nucleotide tokens into 128-dimensional continuous vectors. Three convolutional blocks are then applied, each consisting of a 1D convolution (kernel size 8), batch normalization, ReLU activation, and max pooling (stride 2). 

The kernel size of 8 was selected to capture short sequence motifs spanning approximately 6--8 nucleotides, which aligns with the typical length of many regulatory elements reported in prior genomic studies. The use of three convolutional stages allows the model to progressively expand its receptive field while maintaining computational efficiency. Channel dimensionality is reduced (128~$\rightarrow$~32~$\rightarrow$~16~$\rightarrow$~4) to limit model capacity and reduce overfitting on short sequences.

After flattening, two fully connected layers with dropout ($p{=}0.3$) produce binary class logits.

\subsection{Attention-Based Saliency Computation}

Standard saliency estimation relies on gradient computation 
$\nabla_{\mathbf{x}} f_\theta(\mathbf{x})$ to measure feature importance~\cite{simonyan2013deep}. 
While effective for post-hoc analysis, computing gradients for saliency during training requires additional backward passes and can introduce instability into the optimization process. 

To avoid this overhead, AttnGen estimates importance directly in the forward pass using a lightweight attention mechanism. 
Given the embedding tensor 
$\mathbf{E}(\mathbf{x}) \in \mathbb{R}^{B \times L \times d}$ for a batch of sequences 
($B$: batch size, $L=200$, $d=128$), we compute position-wise scores by averaging across the feature dimension:
\begin{equation}
\mathbf{s}_{b,i}(\mathbf{x}) = \frac{1}{d} \sum_{j=1}^{d} \mathbf{E}_{b,i,j}(\mathbf{x}),
\quad \mathbf{s}(\mathbf{x}) \in \mathbb{R}^{B \times L}.
\end{equation}

The normalized importance weights are then obtained via a softmax operation along the sequence dimension:
\begin{equation}
\mathbf{A}_{b,i}(\mathbf{x}) = 
\frac{\exp\big(\mathbf{s}_{b,i}(\mathbf{x})\big)}
{\sum_{i'=1}^{L} \exp\big(\mathbf{s}_{b,i'}(\mathbf{x})\big)},
\quad \mathbf{A}(\mathbf{x}) \in \mathbb{R}^{B \times L}.
\end{equation}

We use mean aggregation rather than max pooling to reduce sensitivity to individual embedding dimensions, which may be noisy or poorly calibrated during early training. This yields a more stable estimate of position-wise importance.

The resulting weights $\mathbf{A}_{b,i}(\mathbf{x})$ are used to rank nucleotide positions for subsequent masking.

\subsection{Progressive Masking Strategy}

During training, AttnGen uses the attention weights to suppress low-importance positions. 
For a masking ratio $\alpha \in [0,1]$, the number of masked positions is:
\begin{equation}
k = \left\lfloor \alpha L \right\rfloor.
\end{equation}

The indices corresponding to the $k$ least salient positions are defined as:
\begin{equation}
\mathcal{I}_{\text{mask}}(\mathbf{x}) = 
\operatorname{arg\,min}_{\substack{\mathcal{I} \subset \{1,\dots,L\} \\ |\mathcal{I}| = k}} 
\sum_{i \in \mathcal{I}} \mathbf{A}_{b,i}(\mathbf{x}),
\end{equation}

i.e., the set of $k$ positions with the smallest importance scores. The masked input $\tilde{\mathbf{x}}$ is constructed by replacing positions in $\mathcal{I}_{\text{mask}}(\mathbf{x})$ with a padding token (index 0). This operation is applied independently to each sequence in the batch. This masking strategy introduces a controlled perturbation that forces the model to redistribute attention toward more informative regions. 
We consider four masking regimes: baseline (0\%), moderate (10--25\%), high (50\%), and extreme (75\%). 
These regimes span the range from minimal perturbation to near-complete information removal, with finer resolution in the moderate range where the accuracy–interpretability trade-off is most sensitive.

\subsection{Saliency-Guided Loss Function}

The AttnGen objective combines standard classification loss with a consistency constraint:
\[
\mathcal{L}_{\text{total}} = \mathcal{L}_{\text{CE}}\big(f_\theta(\mathbf{x}), y\big) + \lambda\, \mathcal{D}_{\text{KL}}\big(f_\theta(\mathbf{x}) \parallel f_\theta(\tilde{\mathbf{x}})\big),
\]
where $\mathcal{L}_{\text{CE}}$ is the cross-entropy loss and $\mathcal{D}_{\text{KL}}$ measures divergence between prediction distributions:
\[
\mathcal{D}_{\text{KL}}\big(f_\theta(\mathbf{x}) \parallel f_\theta(\tilde{\mathbf{x}})\big) = \sum_{c=1}^C P(c|\mathbf{x}) \log \frac{P(c|\mathbf{x})}{P(c|\tilde{\mathbf{x}})}.
\]

The KL term penalizes large prediction shifts when low-importance positions are removed, encouraging the model to rely on features that remain stable under masking. The regularization weight $\lambda{=}0.1$ controls the trade-off between predictive accuracy and consistency, and was selected based on a small grid search over $\{0.01, 0.1, 0.5\}$ on the validation set. For the baseline setting ($\alpha{=}0$), $\lambda{=}0$ reduces the objective to standard cross-entropy training.

Algorithm~\ref{alg:attngen} summarizes the overall training procedure, replacing gradient-based saliency computation in~\cite{ismail2021improving} with forward-pass attention estimation.

\begin{algorithm}[h]
\footnotesize
\SetAlgoNlRelativeSize{-1}
\caption{AttnGen}
\label{alg:attngen}
\KwIn{Training samples $\mathbf{X}$, labels $\mathbf{y}$, masking ratio $\alpha$, learning rate $\eta$, KL weight $\lambda$, epochs $N$}

\For{$epoch = 1$ \textbf{to} $N$}{
    \For{each mini-batch $(\mathbf{x}, y)$}{
    
        \textbf{\# Compute position-wise importance}\\
        $\mathbf{s}_{b,i} = \frac{1}{d}\sum_{j}\mathbf{E}_{b,i,j}$,\;
        $\mathbf{A}_{b,i} = \frac{\exp(\mathbf{s}_{b,i})}{\sum_{i'}\exp(\mathbf{s}_{b,i'})}$\;
        
        \textbf{\# Select low-importance positions}\\
        $k = \lfloor \alpha L \rfloor$\;
        $\mathcal{I}_{\text{mask}} = \arg\min_{|\mathcal{I}|=k}\sum_{i\in\mathcal{I}}\mathbf{A}_{b,i}$\;
        
        \textbf{Construct masked input}\\
        $\widetilde{\mathbf{x}} = \mathrm{Mask}(\mathbf{x}, \mathcal{I}_{\text{mask}})$\;
        
        \textbf{\# Compute training objective}\\
        $\mathcal{L} = \mathcal{L}_{\text{CE}} + \lambda\,\mathcal{D}_{\text{KL}}$\;
        
        \textbf{\# update}\\
        $\theta \leftarrow \theta - \eta \nabla_\theta \mathcal{L}$\;
    }
}
\end{algorithm}

\subsection{Training Configuration}

All models are trained using the Adam optimizer (learning rate $\eta{=}0.001$), batch size 64, and weight decay $10^{-4}$. These hyperparameters follow standard configurations for short-sequence classification tasks and were found to provide stable convergence in preliminary experiments. 

Early stopping with patience 10 (based on validation accuracy) is applied to balance training time and overfitting risk. The KL weight $\lambda{=}0.1$ is used for all masking levels except the baseline ($\alpha{=}0$). We use a fixed random seed (42), mini-batch shuffling, and gradient clipping (max-norm 1.0) to stabilize optimization.

\subsection{Gradient-Based Importance Validation}

To evaluate whether attention based importance aligns with gradient sensitivity, we perform post-hoc gradient analysis. For a trained model $f_\theta$, we sample $N{=}3000$ sequences and compute importance scores:
\[
I_i = \left\| \frac{\partial f_\theta(\mathbf{x})}{\partial \mathbf{E}(\mathbf{x})_i} \right\|_2,
\]
where $\mathbf{E}(\mathbf{x})_i$ denotes the embedding at position $i$.

We then progressively mask positions in decreasing order of $I_i$ and measure classification accuracy as a function of the number of masked nucleotides $m$. A steeper accuracy drop under high-importance masking—compared to random or low-importance baselines—indicates that the model relies on a structured rather than diffuse set of positio\section{Results}

We evaluate \textbf{AttnGen} on the \texttt{demo\_human\_or\_worm} benchmark~\cite{gresova2023genomic} under multiple masking configurations (10\%, 20\%, 50\%, 75\%). 
All experiments use identical architectures, hyperparameters, and random seeds.

\subsection{Classification Performance}

Table~\ref{tab:main_results} reports classification accuracy across masking levels. 
\textbf{AttnGen(10\%)} achieves 96.73\% accuracy, compared to 95.83\% for the baseline CNN (+0.90~pp). 
\textbf{AttnGen(20\%)} reaches 96.10\%. 

At higher masking levels (50\% and 75\%), accuracy decreases to 95.55\% and 79.81\%, respectively. 
This drop suggests that removing too many positions eliminates information required for classification.

\begin{table}[h]
\centering
\caption{Classification accuracy on the \texttt{demo\_human\_or\_worm} benchmark.}
\label{tab:main_results}
\resizebox{\columnwidth}{!}{
\begin{tabular}{lccc}
\toprule
\textbf{Model} & \textbf{Accuracy (\%)} & \textbf{vs.\ Baseline} & \textbf{Convergence} \\
\midrule
Traditional Genomic CNN & 95.83 & -- & -- \\
AttnGen(10\%) & \textbf{96.73} & +0.90 & faster \\
AttnGen(20\%) & 96.10 & +0.27 & faster \\
AttnGen(50\%) & 95.55 & $-$0.28 & faster \\
AttnGen(75\%) & 79.81 & $-$4.02 & unstable \\
\bottomrule
\end{tabular}}
\end{table}

\subsection{Attention-Based Masking Visualization}

\begin{figure*}[ht]
\centering
\includegraphics[width=1.65\columnwidth]{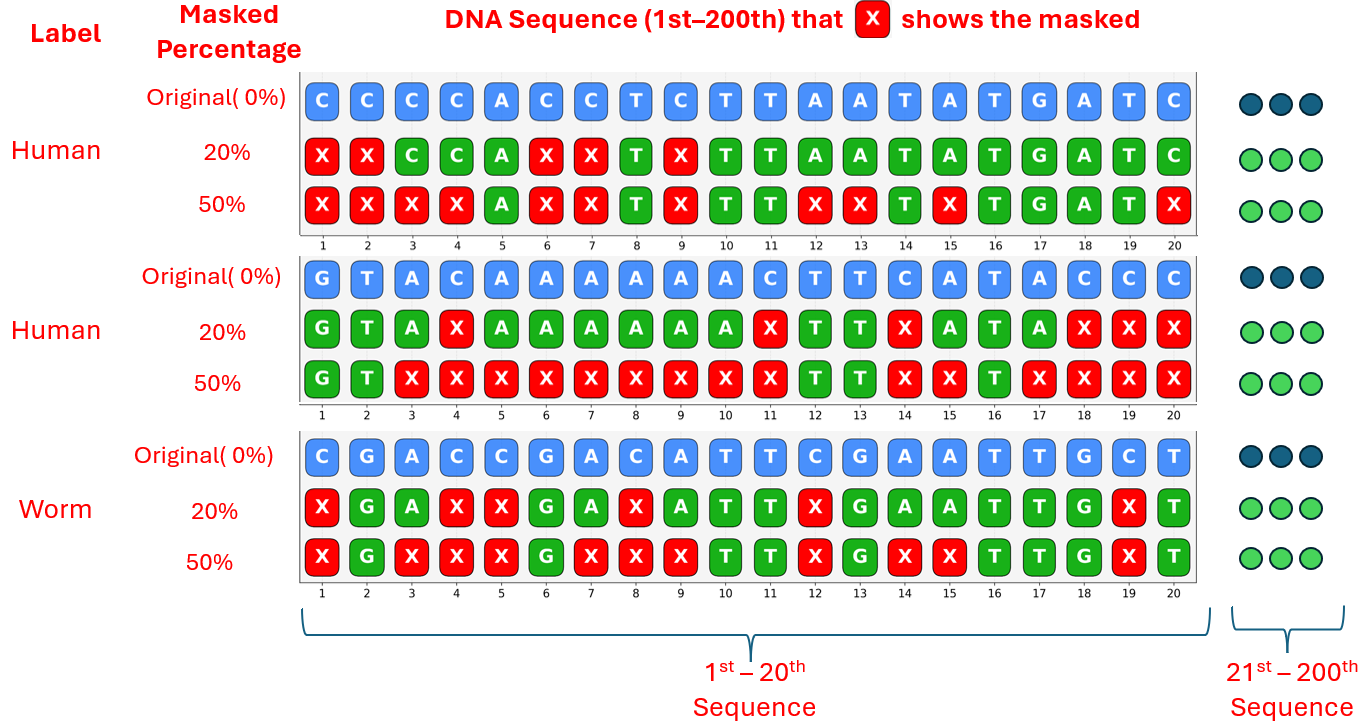}
\caption{Masking patterns for human and worm sequences at different masking levels. Blue: retained positions; red: masked positions.}
\label{fig:saliency_viz}
\end{figure*}

Figure~\ref{fig:saliency_viz} shows masking patterns at different levels. 
At 20\% masking, removed positions are distributed across sequence, while retained positions form compact regions that differ across samples. This variation indicates that masking is driven by sequence specific importance estimates rather than fixed positional patterns. 
At higher masking levels (50\%), retained regions become smaller and less connected, and classification becomes more sensitive to the remaining context.

\subsection{Gradient-Based Importance Analysis}

We compare attention based importance with gradient based rankings by measuring how accuracy changes when high gradient positions are removed. 
For each sequence, position are ranked by gradient magnitude and progressively masked.

\begin{table}[ht]
\centering
\caption{Accuracy under progressive masking of high-gradient positions.}
\label{tab:accuracy_drop}
\begin{tabular}{cccc}
\toprule
\textbf{Masked Positions} & \textbf{Mean Accuracy} & \textbf{Std Dev} & \textbf{Drop} \\
\midrule
0 & 96.91 & $\pm$17.33 & -- \\
1 & 95.17 & $\pm$21.45 & 1.74 \\
5 & 88.40 & $\pm$32.03 & 8.51 \\
10 & 82.43 & $\pm$38.06 & 14.48 \\
25 & 77.67 & $\pm$41.65 & 19.24 \\
50 & 72.17 & $\pm$44.83 & 24.74 \\
100 & 60.10 & $\pm$48.98 & 36.81 \\
150 & 50.83 & $\pm$50.00 & 46.08 \\
200 & 50.00 & $\pm$50.01 & 46.91 \\
\bottomrule
\end{tabular}
\end{table}

Accuracy decreases as more high-gradient positions are removed. 
Masking 10 positions reduces accuracy by approximately 14.5~pp, while masking all positions reduces performance to chance level. The standard deviation increases with the number of masked positions. 
This indicates variability across sequences: some retain partial predictive signal after masking, while others do not. 
This variation is consistent with differences in how informative regions are distributed across inputs.

\begin{figure}[ht]
\centering
\includegraphics[width=1.12\columnwidth]{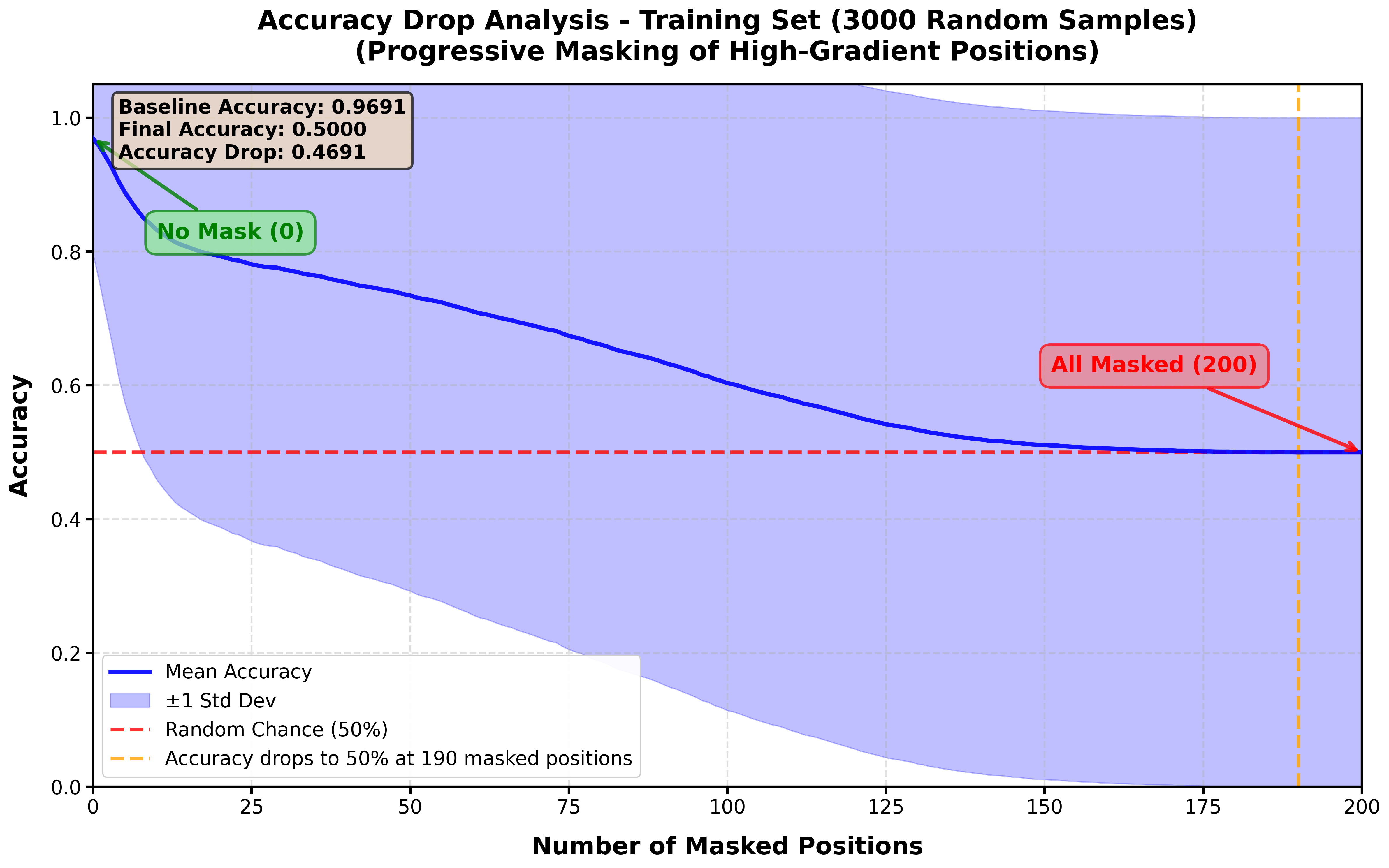}
\caption{Accuracy under progressive masking of high-gradient positions. Shaded region: $\pm$1 std.}
\label{fig:accuracy_drop}
\end{figure}

\subsection{Ablation Study}

We study the 10\% masking configuration to isolate the effects of \textbf{attention-based masking} and \textbf{KL-consistency regularization}. 
Table~\ref{tab:ablation} reports classification accuracy for each variant.

\begin{table}[ht]
\centering
\caption{Ablation on AttnGen(10\%).}
\label{tab:ablation}
\begin{tabular}{lc}
\toprule
\textbf{Configuration} & \textbf{Accuracy (\%)} \\
\midrule
Full Method (Attention + KL) & \textbf{96.73} \\
Random Masking + KL & 95.88 \\
Attention Masking (No KL) & 95.98 \\
Baseline (No Masking) & 95.83 \\
\bottomrule
\end{tabular}
\end{table}

Random masking with KL regularization (95.88\%) performs similarly to the baseline (95.83\%), indicating that KL consistency alone does not improve performance when masking is not guided by importance scores. In contrast, attention-based masking without KL (95.98\%) yields a larger improvement, suggesting that identifying and removing low-importance positions contributes more directly to performance gains. Combining attention-based masking with KL regularization produces the highest accuracy (96.73\%), indicating that attention determines which positions are removed, while KL regularization stabilizes the output distribution when they are.

\subsection{Training Stability}

Moderate masking ratios (10--20\%) produce stable optimization, with validation loss decreasing smoothly and remaining close to training loss throughout training. The 50\% configuration remains stable but shows a wider generalization gap, consistent with the accuracy drop reported in Table~\ref{tab:main_results}.

In contrast, 75\% masking leads to a rapid increase in validation loss after early epochs, indicating that the model fails to maintain generalization when too much input information is removed.

\section{Conclusion}
We presented \textbf{AttnGen}, a training framework that integrates importance estimation into the optimization process for genomic sequence classification. By combining attention-based saliency with progressive masking and KL based consistency, the model focuses on a compact set of informative positions while maintaining predictive performance. On the \texttt{demo\_human\_or\_worm} benchmark, moderate masking (10--20\%) provides the best trade off, reaching up to 96.73\% accuracy. Ablatin results show that attention-guided masking is the primary source of improvement. Additionally, removing high-importance positions leads to a larger drop in accuracy compared to removing low-importance ones, indicating that the learned importance scores align with the model’s predictions. Future work includes extending the method to longer and more complex genomic sequences.

\end{document}